# LLM 幻觉检测：基于隐层时序信号的快速傅里叶变换法

李金欣　涂刚　程圣昱　胡俊杰*

华中科技大学 计算机科学与技术学院, 湖北 武汉 430074

摘　要　尽管大语言模型（Large Language Model, LLM）在语言理解、代码生成等多种任务中展现出强大能力，但是在生成过程中频繁出现的幻觉现象，已成为制约其在关键应用场景中部署的重要障碍。目前主流幻觉检测方法依赖于事实一致性校验或静态隐层特征，前者受限于知识覆盖范围，后者难以捕捉推理过程中出现的偏差。为解决上述问题，本文提出一种新的幻觉检测算法——HSAD（**H**idden **S**ignal **A**nalysis-based **D**etection）。首先，该方法对 LLM 推理过程中的隐层向量进行跨层采样，构建出承载了推理信息的隐层时序信号。接着，使用快速傅里叶变换对隐层时序信号进行频域分析，提取出每个特征维度最大非直流频率分量的模值，从而构造出刻画了 LLM 生成行为的谱特征。最后，从 LLM 的自回归生成机制出发，推导并确定了用于识别模型幻觉现象的最优观测点，从而建立了频域特征提取的完整流程，据此设计实现了一种基于隐层时序信号频域分析的幻觉检测算法。得益于对推理过程建模与频域特征提取的有效结合，HSAD 方法克服了现有方法在知识覆盖和推理偏差检测上的局限，展现出更高的检测精度和鲁棒性。实验结果表明，HSAD 在 TruthfulQA 等多个标准数据集上平均提升超过 10 个百分点，进一步的消融实验验证了频域建模与跨层时序结构等在提升检测性能方面的关键作用。

# LLM Hallucination Detection: A Fast Fourier Transform Method Based on Hidden Layer Temporal Signals

LI Jin-Xin　TU Gang　CHENG Sheng-Yu　HU Jun-Jie

School of Computer Science and Technology, Huazhong University of Science and Technology, Wuhan, HuBei 430074

**Abstract**

Although large language model have demonstrated powerful capabilities in various tasks such as language understanding and code generation, the frequent occurrence of hallucinations during generation has become a significant obstacle to their deployment in critical application scenarios. Current mainstream hallucination detection methods rely on fact consistency checks or static hidden layer features. The former is limited by knowledge coverage, while the latter struggles to capture reasoning biases during the inference process. To address these issues, inspired by psychological signal analysis methods in cognitive neuroscience, we propose a hallucination detection approach based on hidden-layer temporal signal frequency-domain analysis, termed HSAD (**H**idden **S**ignal **A**nalysis-based **D**etection). First, the method constructs a hidden-state temporal signal that encapsulates reasoning information by

 李金欣，男，硕士研究生，主要研究领域为大语言模型、深度学习。E-mail: M202373861@hust.edu.cn。程圣昱，男，硕士研究生，主要研究领域为大语言模型、深度学习。胡俊杰，男，硕士研究生，主要研究领域为大语言模型、深度学习。涂刚（通信作者)，男，博士，副教授，主要研究领域为大语言模型、深度学习、机器学习和自然语言处理。E-mail: tugang@hust.edu.cn。第1 作者手机号码 (投稿时必须提供，以便紧急联系，发表时会删除): 18182224174, E-mail: M202373861@hust.edu.cn

performing cross-layer sampling on the hidden-state vectors during the LLM’s inference process.Next, it applies the Fast Fourier Transform to conduct a frequency-domain analysis of this signal. It then extracts the magnitude of the maximum non-DC frequency component for each dimension, thereby constructing spectral features that characterize the LLM’s generative behavior.Finally, grounded in the LLM’s autoregressive generation mechanism, the approach derives and identifies the optimal observation points for detecting model hallucination. Building upon this foundation, it designs and implements a hallucination detection algorithm based on the frequency-domain analysis of the hidden-state temporal signal. By effectively combining the modeling of the reasoning process with frequency-domain feature extraction, the HSAD method overcomes the limitations of existing methods in knowledge coverage and the detection of reasoning biases, demonstrating higher detection accuracy and robustness. Experimental results show that HSAD achieves an average improvement of over 10 percentage points on multiple standard datasets, including TruthfulQA. Further ablation studies verify the crucial role of frequency-domain modeling and the cross-layer temporal structure in enhancing detection performance.

## 1 引言

LLM 近年来在语言理解、代码生成等任务中展现出卓越性能，成为文本生成、问答系统和信息抽取等应用场景的核心技术基础[1]。然而，其生成过程中频繁出现的幻觉现象[2]，即生成与事实不符或缺乏上下文支持的结果，不仅削弱了 LLM 输出的可信度，也严重制约了其在高风险场景中的部署[3]。

在认知神经科学研究中，大量实验证据表明，人类在面临信息伪造或认知冲突的场景时，会在一定时间尺度上呈现出特定的心理和神经信号变化，例如认知负荷逐步上升、注意力状态波动以及脑电信号在频谱结构上的演化[4]。如图 1所示，这些随时间演化的信号模式可被视为内在冲突的可观测生理表现，能够间接反映个体的认知状态及信息处理的动态过程。通过分析此类信号，可为人类的测谎研究提供理论依据与技术支撑。实践证实，LLM 在生成幻觉内容时，常表现出与人类在信息伪造情境下相似的行为模式。因此，可借鉴认知神经科学中的信号建模思路，将 LLM 的推理过程建模为一种时序信号，为幻觉检测提供新的视角。

HSAD 方法受此启发，将 LLM 推理过程中不同层的隐层向量建模为隐层时序信号，类比人类的测谎过程，对 LLM 进行幻觉检测。

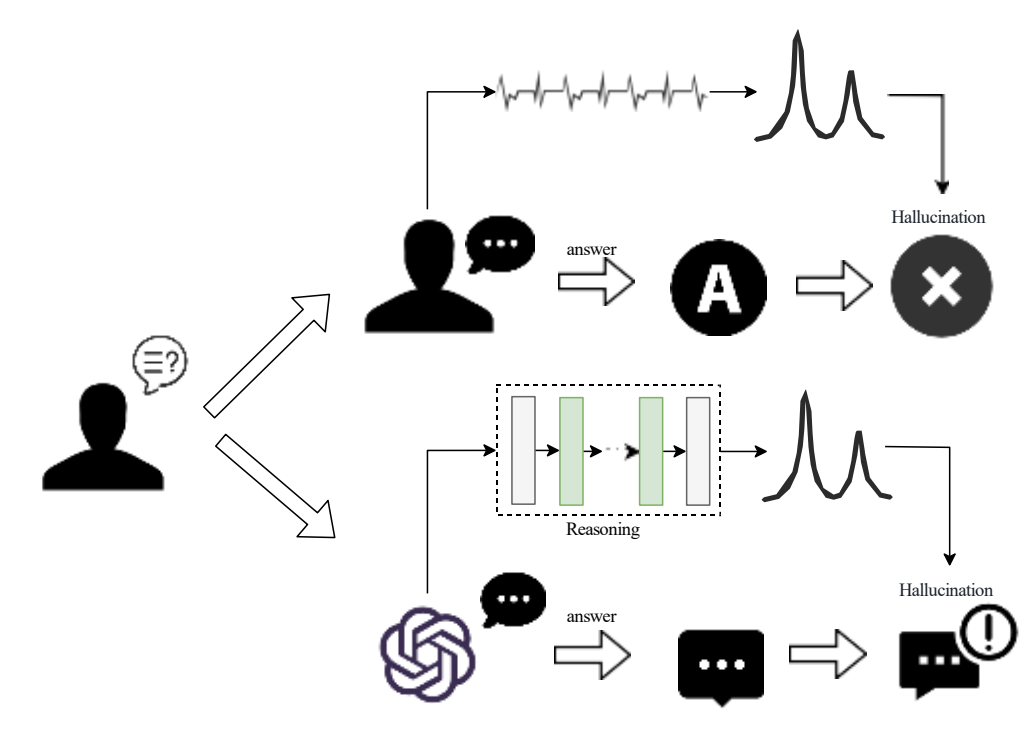

图 1 对人类测谎与对模型测谎对比示意图

具体而言，如图 1所示，首先，HSAD 从 LLM 的前向推理过程中提取隐层向量，按层序构建为具有时间特性的隐层时序信号；随后，借助快速傅里叶变换将该隐层时序信号映射至频域，构造出谱特征，从而揭示其在推理时的异常信号，实验证明，这些异常信号往往与生成内容中的幻觉相对应；在此基础上，进一步将谱特征作为判别依据，设计幻觉检测算法，用于检测 LLM 生成过程中可能存在的幻觉内容。这一策略旨在以可解释的方式刻画 LLM 思维过程的演化轨迹，从而实现对幻觉的有效检测[5-6]。

与现有多数方法不同，HSAD 并不依赖外部知

识库进行事实核验，而是专注于 LLM 内部信号自身的异常性检测。论文主要贡献如下。

① 提出了一种类比人类测谎机制的 LLM 幻觉检测方法 HSAD。基于 LLM 的推理过程构建隐层时序信号，并引入频域分析与谱特征构造，对推理过程中可能存在的幻觉进行分析与识别；理论推导与实验证明了该方法的可行性与有效性。

② 基于谱特征设计并实现了 LLM 幻觉检测算法。首先推导并确定了 LLM 幻觉观测点，然后在频域中构造谱特征用于检测；该算法在多个标准数据集上达到 SOTA，同时通过消融实验验证了各模块的独立贡献与协同增益。

## 2 相关工作

现有幻觉检测方法主要依赖于对 LLM 可解释性的研究，具体可分为两类：（1）基于事实一致性的验证；（2）基于隐层向量的分析，用于识别和定位生成内容中的潜在虚假信息。可解释性研究旨在揭示 LLM 内部机制与决策逻辑，是提升 LLM 可靠性与可控性的关键路径[7]。主流方法包括基于观察（如探针分析、Logit 镜像、稀疏表示、跨模型解释）、基于干预（如激活补丁[8]）、观察与干预融合的方法，以及其他技术路径（如转码器[9]、概念驱动解释[10]、分类器注入、基于奇异值分解的注意力分析等）。这些方法为识别幻觉相关特征提供了理论支撑，推动了对 LLM 语义演化与行为机制的系统性理解。

基于事实一致性的验证是当前幻觉检测方法的重要方向，其核心思想是将生成内容与外部权威知识进行对齐，以识别潜在虚假信息[11]。Min 等人提出的 FActScore [12] 通过将生成文本划分为原子事实单元，并与权威知识库（如维基百科）对齐验证，以评估文本的事实一致性。然而，该方法高度依赖知识库的覆盖率与更新频率，限制了其在开放领域中的适用性。Wei 等人[13] 开发的 SAFT 方法在此基础上引入搜索引擎交互机制，利用 LLM 代理对原子事实进行多轮查询验证，从而增强了长文本生成的事实性评估能力。但该方法计算成本较高，且受限于知识覆盖范围。

隐层向量分析为幻觉检测提供了新的技术路径[14]。Fu 等人提出的 INSIDE 方法[15]，通过解析 LLM 内部密集语义信息，有效检测幻觉，并引入特征剪辑技术，调控隐层激活以缓解生成过程中的过度自信，从而提升幻觉识别能力。Duan 等人[16]通过实证分析表明，LLM 在真实与虚构回答任务中的隐层状态存在显著差异，为基于隐层向量的幻觉检测奠定了基础。Zhou 等人[17]进一步引入傅里叶变换，从频域角度揭示 LLM 对真实与虚构信息处理的差异性频率特征，拓展了对幻觉产生机制的理解视角。 Ju 等[18]人利用探针分析了 LLM 各层对上下文知识的编码能力；Jin 等[19]人则揭示了不同层学习不同概念的能力；He 等[20]人结合 BLiMP 基准，发现随着句子结构复杂度增加，模型需要更深层次来学习语法信息。TTPD[21] 在 LLM 中间层构造“真理子空间”，通过两个正交方向区分真实与谎言，具备对否定句和复杂表达的鲁棒检测能力。TruthX [22] 通过自编码器解耦语义与真实度空间，并在推理阶段沿 “真实方向”编辑中间表示，显著缓解幻觉并增强生成内容真实性。

尽管现有研究在 LLM 输出结果或特定隐层向量的静态分析方面取得了一定进展，但仍存在受限于知识覆盖范围、难以反映推理过程中的动态异常等问题。HSAD 为弥补这些不足，引入频域变换思路。该方法在 LLM 前向推理过程中，提取隐层向量并构建隐层时序信号，然后对该信号各维度进行频域变换，构造谱特征，用以捕捉推理过程中出现的异常。文章接下来将对该方法进行详细介绍。第三章将首先对 LLM 的推理过程进行建模分析，并在此基础上构建具有时间结构的隐层时序信号；第四章将先推导幻觉观测点，然后围绕该时序信号，详细介绍谱特征的构造方法以及基于谱特征的幻觉检测算法；第五章则通过多组实验与消融分析，系统验证所提方法的有效性与鲁棒性。

## 3 建模及分析

为理解 LLM 在文本生成过程中的内部认知机制，本章节基于其前向推理过程建立形式化数学模

型，并给出幻觉判别准则。然后从时序建模的角度出发，将不同层中的隐层向量视为推理行为在时域上的展开，进而为后续的频域建模提供基础。

### 3.1 LLM 推理过程建模

本小节将系统梳理 LLM 各层中隐层向量的计算逻辑及其符号体系。在此基础上，提出推理过程的统一表达，并通过图示展示。

设输入问题为 $Q = (t^Q_0, t^Q_1, \dots, t^Q_{m-1})$，对应的参考答案为 $A = (t^A_0, t^A_1, \dots, t^A_{n-1})$，其中 $t_i$ 表示输入或输出序列中的第 $i$ 个 token。如图 2所示，对于一个拥有 $l$ 层解码器的 LLM，其在每一层的计算可表示如下。

$t_i$ 在第 $j$ 层的注意力向量 $ah^j_i$ 如式 1所示。

$$ah^j_i = ATT\,(h^{j-1}_i, k_{cache}, v_{cache}) \tag{1}$$

$t_i$ 在第 $j$ 层的 MLP 向量 $mh^j_i$ 如式 2所示。

$$mh^j_i = MLP\,(rh^j_i) \tag{2}$$

其中 $rh^j_i$ 是注意力的残差向量。

$t_i$ 在第 $j$ 层的最终隐层向量 $h^j_i$ 如式 3所示。

$$h^j_i = Layer^j(h^{j-1}_i) = (ATT \circ MLP\,)(h^{j-1}_i) \tag{3}$$

上述每一层 $Layer^j(*)$ 由注意力子层与 MLP 子层构成，构成跨层信息传递的主干路径。

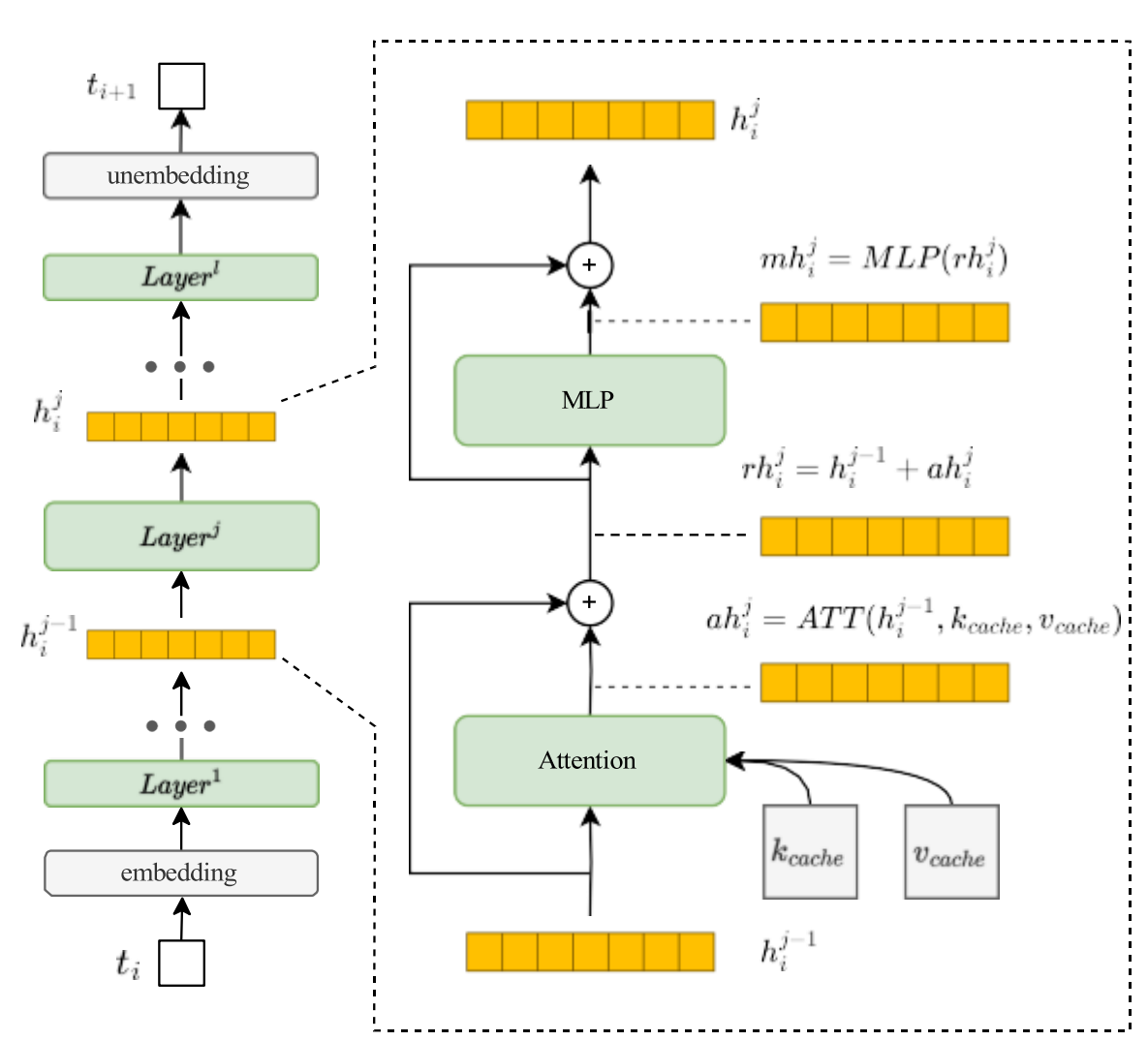

图 2　推理过程 $F_i$ 示意图

将 LLM 对 token $t_0, \dots, t_i$ 的处理定义为第 $i$ 步推理过程 $F_i$，如式 4所示。

$$\begin{aligned} F_i(t_0, \dots, t_i) = (&emb \circ Layer^1 \circ \cdots \circ Layer^l \\ &\circ\, unemb)(t_0, \dots, t_i) \end{aligned} \tag{4}$$

如图 2所示，在 LLM 推理过程中，隐层向量随着推理不断更新、整合语义上下文，具有明显的时序特征和信息流动性。因此，整个推理过程可以类比为一种随时间演进的思想历程。基于此，将不同层的隐层向量构建为隐层时序信号，进行频域建模与分析。

### 3.2 LLM 隐层时序信号构建

定义 LLM 隐层时序信号为 $X \in \mathbb{R}^{4l}$，该信号本质上是隐层向量的跨层采样，用于分析 LLM 内部表示随层深的变化特性。

隐层时序信号具体构建过程如下。

设 LLM 具有 $l$ 层 Transformer 结构，每层包含注意力子层和 MLP 子层，均带有残差连接。从每一层中采样与生成 token $t_{i+1}$ 对应的四个关键节点的向量，将其在第 $j$ 层拼接为一个 $d \times 4$ 的矩阵 $\mathbf{V}^j$，如式 5所示。

$$\mathbf{V}^j = \begin{bmatrix} h^j_i \\ mh^j_i \\ rh^j_i \\ ah^j_i \end{bmatrix}^T \in \mathbb{R}^{d\times 4} \tag{5}$$

四个关键节点的向量分别为：（1）注意力输出向量 $ah^j_i$：反映当前层对上下文依赖关系的建模结果；（2）注意力残差向量 $rh^j_i$：叠加前层信息与注意力输出；（3）MLP 输出向量 $mh^j_i$：体现非线性特征映射后的局部决策依据；（4）第 $j$ 层输出向量 $h^j_i$：该层所有子模块输出的融合结果。

将所有 $l$ 层的 $\mathbf{V}$ 进行拼接，构成矩阵 $\mathbf{T}$，如式 6所示。

$$\mathbf{T} = [\mathbf{V}^l, \dots, \mathbf{V}^j, \dots, \mathbf{V}^1]^T \in \mathbb{R}^{4l\times d} \tag{6}$$

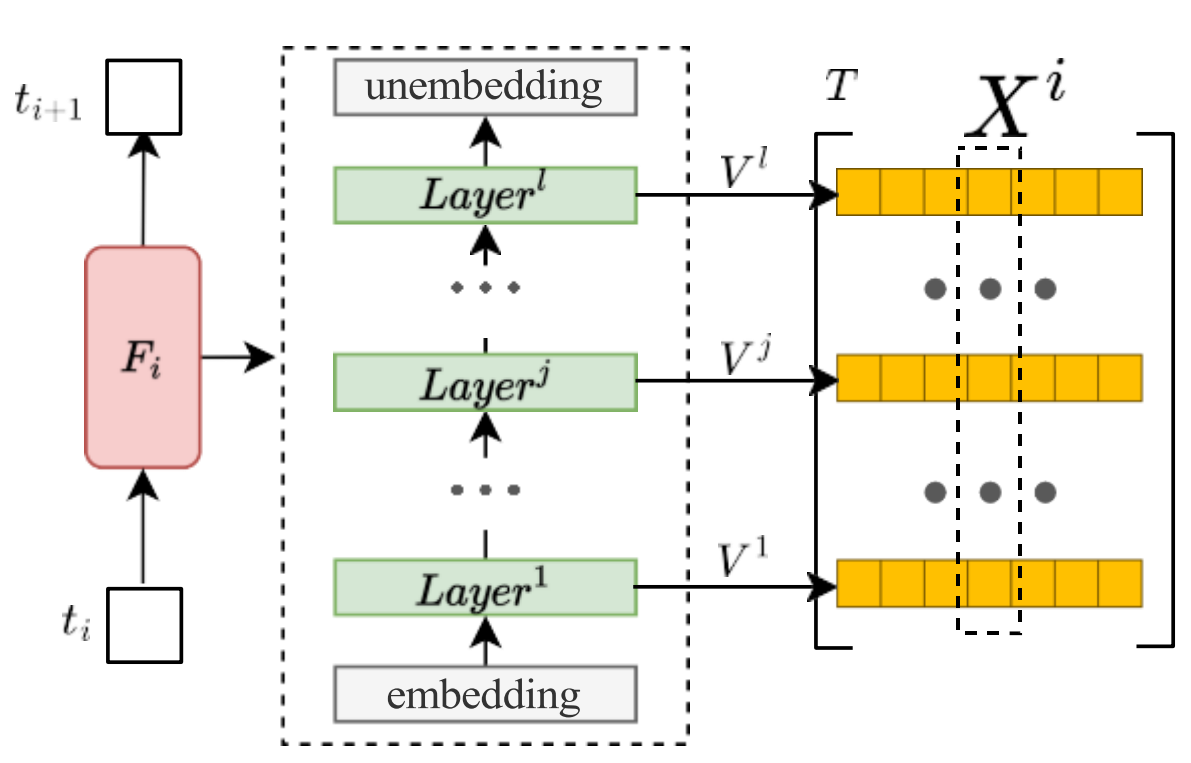

图 3　隐层时序信号构建

如图 3 所示，在矩阵 $T$ 中，每一列对应一个维度随时间的变化过程。对于 $i \in [1, d]$，将矩阵 $T$ 的第 $i$ 列定义为第 $i$ 个维度的隐层时序信号 $X^i$，其刻画了该维度在 LLM 推理过程中随着推理进行而产生的变化。

### 3.3　LLM 幻觉判别

为判别 LLM 输出是否构成幻觉，使用如式 7所示的判别准则。

$$K(A \mid Q) = \begin{cases} 1 & sim(A, A^*) \leq \tau \\ 0, & otherwise \end{cases} \tag{7}$$

其中 $Q$ 为输入问题，$A$ 为 LLM 生成回答，$A^*$ 为参考答案，$sim()$ 表示语义相似度评分，$\tau$ 为设定阈值。满足上述条件的样本视为幻觉输出。

## 4　LLM 幻觉检测：基于隐层时序信号的快速傅里叶变换法

本章节将介绍基于对隐层时序信号进行快速傅里叶变换的幻觉检测法。首先推导并确定了 LLM 的幻觉观测点，然后通过对隐层时序信号进行快速傅里叶变换，构造出谱特征并证明其有效性，最后基于谱特征进行幻觉检测。

### 4.1　LLM 幻觉观测点的确定

在 3.1 节中已形式化定义了 LLM 每一步的推理过程 $F_i$。对于输入问题 $Q$,其回答 $A = (t^A_0, \ldots, t^A_{n-1})$ 是在多个步骤中逐步生成的，完整过程对应于一系列推理状态 $F^A_0, \ldots, F^A_{n-1}$。

若希望基于内部信号检测幻觉，需分析这些推理状态中的隐层时序信号，但逐步处理将带来较高计算成本。因此，引出一个关键问题：是否必须分析所有推理状态，或仅选取一个代表性节点即可。

为此，从生成结构出发，提出并证明了以下命题：最终步推理状过程 $F^A_{n-1}$ 已整合了从输入 $Q$ 到输出 $A$ 的全部上下文信息，可作为整体推理行为的观测点。

**定理 1.** 在标准自回归结构中，最终步推理 $F^A_{n-1}$ 蕴涵从输入 $Q$ 到输出 $A$ 的全部推理过程信息。

$$F^A_{n-1} \supseteq \left\{ F^Q_0, \ldots, F^Q_{m-1}, F^A_0, \ldots, F^A_{n-2} \right\} \tag{8}$$

证明.　考虑 $F^A_{n-1}$ 的输入序列

$$Input_{n-1} = (t^Q_0, \ldots, t^Q_{m-1}, t^A_0, \ldots, t^A_{n-2}) \tag{9}$$

根据前向传播定义

$$F^A_{n-1} = f(t^Q_0, \ldots, t^Q_{m-1}, t^A_0, \ldots, t^A_{n-2}) \tag{10}$$

考虑自注意力结构，每一层 $j$ 对历史 token 应用注意力

$$ATT(t_i) = \sum_{k=0} \alpha^j_{ik} \cdot V^j_k \tag{11}$$

其中 $\alpha^j_{ik}$ 为注意力权重，$V^j_k$ 为第 $j$ 层第 $k$ 个 token 的 value 表示。显然，$F_i$ 包含 $F_0, \ldots, F_{i-1}$ 的信息

$$F_i \supseteq F_0, F_1, \ldots, F_{i-1} \tag{12}$$

因此

$$F^A_{n-1} \supseteq \{F^Q_0, \ldots, F^Q_{m-1}, F^A_0, \ldots, F^A_{n-2}\} \tag{13}$$

证毕.

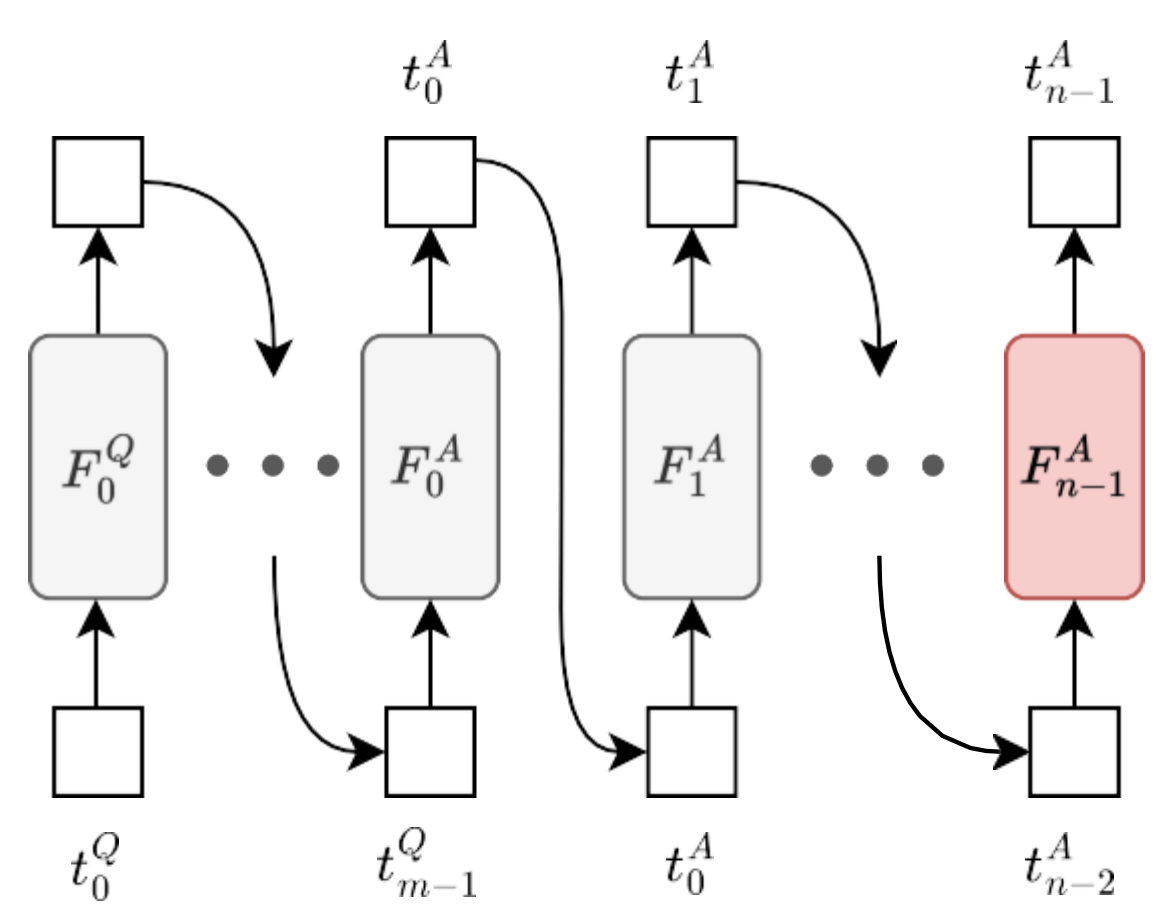

图 4 最终推理过程 $F_{n-1}^A$ 蕴涵全局推理信息

综上，如图 4所示，$F_{n-1}^A$ 可被视为 LLM 输入-输出路径中的“全局推理蕴涵”，其隐层向量集中体现了 LLM 对问题 $Q$ 的理解与对回答 $A$ 的组织能力。

通俗而言，$F_{n-1}^A$ 表示模型在生成最后一个token 时的推理状态，此时它已接收完整的问题 $Q$ 及此前生成的全部内容，是对当前任务理解最为充分的时刻。该节点不仅集中体现了语义一致性与因果逻辑，也是观测 LLM 是否存在幻觉的理想观测点。

**4.2** 基于隐层时序信号的频域变换与谱特征构造

在信号处理领域，对信号进行分析时，往往需要同时考察时域与频域特征。HSAD 通过快速傅里叶变换对每一维度的时间信号进行频谱分析。

首先，对矩阵 **T** 的每一列隐层时序信号 $X^i$ 执行如式 14所示的快速傅里叶变换。

$$Y_k^i = \sum_{n=0}^{N-1} X_n^i \cdot W_N^{nk} \qquad (14)$$

其中，$Y_k^i$ 为变换后得到的第 $k$ 个频率分量的复数幅值，$X^i{}_n$表示 **T** 中第 $i$ 列第 $n$ 个元素，$N = 4l$，旋转因子如式 15所示。

$$W_N^{nk} = e^{-j\frac{2\pi}{N}nk} \qquad (15)$$

该因子将时间域信号映射到频域中对应频率的正余弦基上，从而揭示其在不同频率下的成分。

实践证明，频域中直流分量通常对应输入信号的整体偏移量，难以反映推理过程中的变化，因此在特征提取过程中对其予以剔除。

具体而言，仅保留最大非直流频率分量的模值，用以捕捉 LLM 推理过程中的时序异常信息，如式 16所示。

$$A^i = \max_{1\le k<N} |Y_k^i|, \quad i = 1, 2, \ldots, d \qquad (16)$$

最终构造出频域表示向量 $f$，如式 17所示。

$$f = [A^1, A^2, \ldots, A^d] \in \mathrm{R}^d \qquad (17)$$

如图 5所示，该频域向量 $f$ 即为输入 $Q$ 的谱特征，能够有效刻画 LLM 在推理过程中的异常扰动，为幻觉检测提供结构化输入。

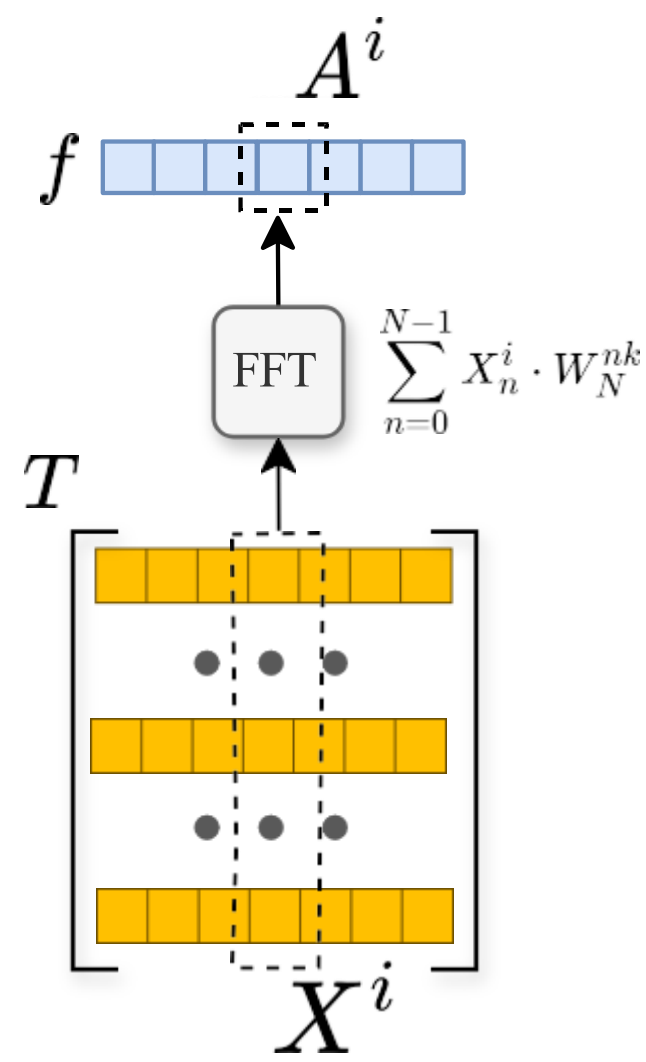

图 5 基于构建好的隐层时序信号借助快速傅里叶变换进行谱特征的构造；该谱特征可作为幻觉检测等下游任务的结构化输入。

为了验证所构造的谱特征 $f$ 是否具备实际意义和对 LLM 生成行为的刻画能力，设计并开展了一系列可控实验。以其中一个典型实验为例，在 LLM 生成过程中，通过屏蔽掉某些频域幅度较大的维度，观察其输出的变化。

如图 6 所示，在保持原始上下文不变的前提下，一旦屏蔽掉某些对应高频幅度分量的隐层维度，LLM 的回答会发生显著且规律的稳定变化。例如，无法正确回答原本能够回答的问题或原本以中文为

主的回答转变为英文输出。这种变化不仅明显，且在多次实验中展现出一致性和可复现性。

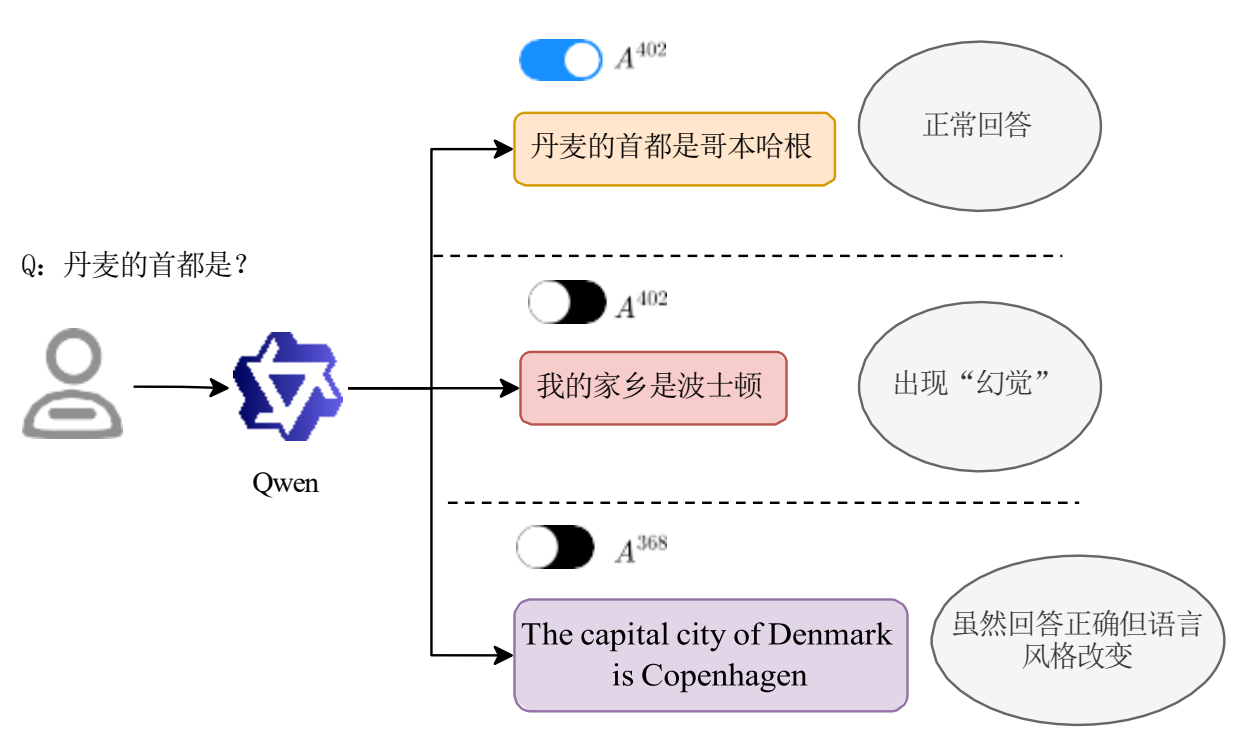

图 6 在保持原始上下文不变的前提下，屏蔽掉某些高频幅度分量对应的维度，对 LLM 回答的影响

一系列实验的结果进一步验证了频域建模的有效性。某些幅度较大的频域特征维度在一定程度上编码了语言风格、语义一致性与推理等关键信息。谱特征 $f$ 在生成中发挥关键作用，将其作为结构化输入用于幻觉检测等下游任务是合理且有效的。

### 4.3 基于谱特征的幻觉检测

基于构造出的谱特征，构建幻觉检测器 $H: \mathbb{R}^d \rightarrow \{0, 1\}$，输入为谱特征 $f$，输出为幻觉检测结果。

对于问题集合 Ques$(Q, A^*)$，$\forall Q$，幻觉检测器 $H$ 的目标是学习为一个二元预测器，如式 18所示。

$$H(Q) = \begin{cases} 1 & if \ K(A|Q) = 1 \\ 0, & otherwise \end{cases} \tag{18}$$

其中 $K(A|Q)$ 表示答案 $A$ 是否存在幻觉。

为了在保持较强非线性建模能力的同时兼顾部署效率，本研究在频谱特征分类阶段采用一种改进型多层感知机（Enhanced MLP）作为检测器。该网络结构包含特征提取模块与分类模块两个部分：特征提取模块由多层全连接层、批归一化、ReLU 激活及Dropout 组成，逐层将特征维度压缩至 256；分类模块为单层全连接层，将 256 维表示映射至二元分类输出空间 $\mathbb{R}^2$。

其核心计算过程可描述为

$$h = ReLU\left(BN(Wf + b)\right) \tag{19}$$

$$\hat{y} = \sigma(Wh + b) \tag{20}$$

其中 $BN()$ 表示批归一化操作，$\sigma()$ 为 Sigmoid 激活函数，输出幻觉概率 $\hat{y} \in (0, 1)$。这种分阶段降维与正则化结合的设计，可以在不显著增加参数量的情况下捕获复杂频谱模式，并有效抑制过拟合。

与传统分类方式相比，逻辑回归虽结构简洁，但对非线性频率特征的区分能力有限，准确率存在瓶颈；SVM 在小样本条件下具有一定优势，尤其是线性 SVM 在频谱特征较稀疏时收敛较快；但在跨任务域或跨模型推理场景下，其泛化稳定性不足；Enhanced MLP 在泛化能力与表现稳定性方面均优于上述方法，尤其在频域特征分布不稳定的场景中具有更强适应能力；此外，该网络可端到端训练，便于集成到更大规模的对齐优化框架中。

模型训练过程中采用二元交叉熵损失函数，并结合L1 正则化以控制谱通道激活的稀疏性，目标函数如式 21所示：

$$\begin{aligned} L_{halluc} = & -[y \log \hat{y} + (1 - y) \log(1 - \hat{y})] \\ & + \lambda \|W_1\|_1 \end{aligned} \tag{21}$$

其中第一项鼓励模型收敛至合理的判别边界，第二项 L1 正则项提升特征选择性，弱化非判别性频率的影响，从而增强泛化能力。

由于频谱特征 $f$ 的提取可在后端并行完成，整体检测流程仅占用极少计算资源。

## 5 实验

在本章节中，首先介绍了实验的环境设置，并展示了 HSAD 相较于其他幻觉检测方式在多个数据集上的优势。然后，通过详尽的消融实验证明频域建模、跨层结构、关键节点融合与合适的幻觉观测点的作用。

### 5.1 实验环境

本小节从数据集、评估模型、评价指标、基线、实现细节五个方面阐述实验环境的设置。

#### 5.1.1　数据集

实验选取了四个生成式问答（QA）任务进行评估。三个开放域 QA 数据集：TruthfulQA、TriviaQA 和 NQ Open；以及特定领域的 QA 数据集 SciQ。选取的问答对数量如表 1所示。将每个数据集的 30% 数据用于测试，其余数据则用于训练。在数据预处理阶段，遵循标准流程对问答对进行清洗和格式化。

表 **1**　数据集

| 数据集名称 | 问答对数量 |
|---|---|
| TruthfulQA | 817 |
| SciQ | 1,000 |
| NQ Open | 3,610 |
| TriviaQA | 9,960 |

#### 5.1.2　评估模型

如表 2所示，使用两类模型进行评估：LLaMA-3.1-8B 和 Qwen-2.5-7B-instruct。

表 **2**　**LLaMA-3.1-8B** 与 **Qwen-2.5-7B-instruct** 模型结构对比

| 特征项 | **LLaMA-3.1-8B** | **Qwen-2.5-7B-instruct** |
|---|---|---|
| 总参数量 | 8.0B | 7.61B |
| 层数 | 32 | 28 |
| 隐层维度 | 4096 | 3584 |
| 注意力机制 | GQA | GQA |
| 中间层维度 | 14336 | 18944 |
| 激活函数 | SwiGLU | SwiGLU |
| 上下文长度 | 128k | 128k |
| 位置编码 | Rope | Rope |
| 归一化 | RMSNorm | RMSNorm |
| 发布机构 | Meta AI | Alibaba |

#### 5.1.3　评价指标

在实验时，使用 ACC 和 AUROC 作为主要评估指标。

ACC 指的是准确率，其计算公式如式 22所示。

$$ACC = \frac{TP + TN}{TP + FP + FN + TN} \tag{22}$$

AUROC 则用于衡量二分类模型在不同阈值下的性能，代表在一个正负样本对中，模型预测正样本的概率高于负样本的概率。其计算如式23所示，$M$ 和 $N$ 代表正样本和负样本的个数，CorrectPair 代表模型预测正样本的概率高于负样本的正负样本对的个数。

$$AUROC = \frac{CorrectPair}{M \times N} \tag{23}$$

#### 5.1.4　基线

如下表所示，将本研究的方法与一系列全面的基线方法进行了比较，其中包括现有的最新方法[23]。这些方法依据各自所基于的理论分为以下四类。

表 **3**　基准线方法分类

| 方法分类 | 方法名称 |
|---|---|
| 基于不确定性 | Perplexity<br>LN-entropy<br>Semantic Entropy (SE) |
| 基于一致性 | Lexical Similarity (LS)<br>SelfCKGPT<br>EigenScore |
| 基于提示 | Verbalize<br>Self-evaluation (Seval) |
| 基于内部状态 | CCS<br>HaloScope |

为了保证公平比较，所有方法均在相同的测试数据集上进行评估，并采用其各自文献中指定的默认实验配置。

#### 5.1.5　实现细节

在生成任务中，对两类模型采样相同的设置：参数 top_p 为 1.0，temperature 参数为 1.0，top_k 为 50，max_new_tokens 为 64。使用 beam search 检测多条潜在回答路径，num_beams 参数为 5。在训练 H 时使用 Adam 优化器训练 50 个 epoch，初始学习率为 5e-4，余弦学习率衰减，batch size 为 128，而权重衰减为 1e-4。

### 5.2 主要结果

为验证所提出 HSAD 方法的有效性与泛化能力，在TruthfulQA、TriviaQA、SciQ 及 NQ Open 四个公开数据集上开展幻觉检测实验，并与多种现有检测方法进行了系统对比。实验中，将BLEURT 分数[24] 作为幻觉判定阈值，并采用 AUROC 作为性能评价指标。

表 4给出了在不同基线模型下各方法的检测性能。可以看出，在所测试的四个数据集上，HSAD 均取得了最高的 AUROC 值，相较于现有方法在多项指标上均有显著提升。例如，在 Qwen-2.5-7B-instruct模型上，HSAD 在 TruthfulQA、TriviaQA、SciQ 和 NQ Open 数据集上的AUROC 分别达到 82.5%、92.1%、 94.7% 和 88.3%，相比性能次优的方法平均提升超过 10 个百分点。在 LLaMA-3.1-8B 模型上，HSAD同样在全部数据集上取得最优结果，进一步验证了该方法在不同模型和数据分布下的稳定性与泛化能力。

**表 4 各方法在不同数据集上的幻觉检测结果（AUROC，%）**

| 模型 | 方法 | TruthfulQA | TriviaQA | SciQ | NQ Open |
|---|---|---|---|---|---|
| Qwen-2.5-7B-instruct | Perplexity | 65.1 | 50.2 | 53.4 | 51.2 |
| | LN-entropy | 66.7 | 51.1 | 52.4 | 54.3 |
| | SE | 66.1 | 58.7 | 65.9 | 65.3 |
| | LS | 49.0 | 63.1 | 62.2 | 61.2 |
| | SelfCKGPT | 61.7 | 61.3 | 58.6 | 63.4 |
| | Verbalize | 60.0 | 54.3 | 51.2 | 51.2 |
| | EigenScore | 53.7 | 62.3 | 63.2 | 57.4 |
| | Self-evaluation | 73.7 | 50.9 | 53.8 | 52.4 |
| | CCS | 67.9 | 53.0 | 51.9 | 51.2 |
| | HaloScope | 81.3 | 73.4 | 76.6 | 65.7 |
| | **HSAD** | **82.5** | **92.1** | **94.7** | **88.3** |
| LLaMA-3.1-8B | Perplexity | 71.4 | 76.3 | 52.6 | 50.3 |
| | LN-entropy | 62.5 | 55.8 | 57.6 | 52.7 |
| | SE | 59.4 | 68.7 | 68.2 | 60.7 |
| | LS | 49.1 | 71.0 | 61.0 | 60.9 |
| | SelfCKGPT | 57.0 | 80.2 | 67.9 | 60.0 |
| | Verbalize | 50.4 | 51.1 | 53.4 | 50.7 |
| | EigenScore | 45.3 | 69.1 | 59.6 | 56.7 |
| | Self-evaluation | 67.8 | 50.9 | 54.6 | 52.2 |
| | CCS | 66.4 | 60.1 | 77.1 | 62.6 |
| | HaloScope | 70.6 | 76.2 | 76.1 | 62.7 |
| | **HSAD** | **81.5** | **86.7** | **85.5** | **80.7** |

综上，HSAD 在不同数据集与基线模型下均取得了优异且稳定的检测效果，充分证明了该方法在幻觉检测任务中的实用性与跨域泛化能力。

## 参 考 文 献

[1] TOUVRON H, MARTIN L, STONE K, et al. Llama 2: Open foundation and fine-tuned chat models[J/OL]. CoRR, 2023, abs/2307.09288. https://doi.org/10.48550/arXiv.2307.09288. DOI: 10.48550/ARXIV.2307.09288.

[2] I Z, LEE N, FRIESKE R, et al. Survey of hallucination in natural language generation[J/OL]. ACM Comput. Surv., 2023, 55(12): 248:1-248:38. https://doi.org/10.1145/3571730.

[3] LI J, CHEN J, REN R, et al. The dawn after the dark: An empirical study on factuality hallucination in large language models[C/OL]//KU L W, MARTINS A, SRIKUMAR V. Proceedings of the 62nd Annual Meeting of the Association for Computational Linguistics (Volume 1: Long Papers). Bangkok, Thailand: Association for Computational Linguistics, 2024: 10879-10899. https://aclanthology.org/2024.acl-long.586/. DOI: 10.18653/v1/2024.acl-long.586.

[4] LO Y H, TSENG P. Electrophysiological markers of working memory usage as an index for truth-based lies [J/OL]. Cognitive, Affective, & Behavioral Neuroscience, 2018, 18(6): 1089-1104. https://pubmed.ncbi.nlm.nih.gov/30022430/. DOI: 10.3758/s13415-018-0624-2.

[5] LIEBERUM T, RAJAMANOHARAN S, CONMY A, et al. Gemma scope: Open sparse autoencoders everywhere all at once on gemma 2[J/OL]. CoRR, 2024, abs/2408.05147. https://doi.org/10.48550/arXiv.2408.05147. DOI: 10.48550/ARXIV.2408.05147.

[6] LINDSEY J, GURNEE W, AMEISEN E, et al. On the Biology of a Large Language Model[Z]. 2025.

[7] JI Z, GU Y, ZHANG W, et al. ANAH: Analytical annotation of hallucinations in large language models [C/OL]//KU L W, MARTINS A, SRIKUMAR V. Proceedings of the 62nd Annual Meeting of the Association for Computational Linguistics (Volume 1: Long Papers). Bangkok, Thailand: Association for Computational Linguistics, 2024: 8135-8158. https://aclanthology.org/2024.acl-long.442/. DOI: 10.18653/v1/2024.acl-long.442.

[8] ZHAO C, QIAN W, SHI Y, et al. Automated natural language explanation of deep visual neurons with

large models (student abstract)[C/OL]//WOOLDRIDGE M J, DY J G, NATARAJAN S. Thirty-Eighth AAAI Conference on Artificial Intelligence, AAAI 2024, Thirty- Sixth Conference on Innovative Applications of Artificial Intelligence, IAAI 2024, Fourteenth Symposium on

Educational Advances in Artificial Intelligence, EAAI 2014, February 20-27, 2024, Vancouver, Canada. AAAI Press, 2024: 23712-23713. https://doi.org/10.1609/aaai.v38i21.30537. DOI: 10.1609/AAAI.V38I21.30537.

[9] TURPIN M, MICHAEL J, PEREZ E, et al. Language models don’t always say what they think: Unfaithful explanations in chain-of-thought prompting[C/OL]//Thirty-seventh Conference on Neural Information Processing Systems. 2023. https://openreview.net/forum?id=bzs4uPLXvi.

[10] WAN D, SINHA K, IYER S, et al. ACUEval: Fine-grained hallucination evaluation and correction for abstractive summarization[C/OL]//KU L W, MARTINS A, SRIKUMAR V. Findings of the Association for Computational Linguistics: ACL 2024. Bangkok, Thailand: Association for Computational Linguistics, 2024: 10036-10056. https://aclanthology.org/2024.findin gs-acl.597/. DOI: 10.18653/v1/2024.findings-acl.597.

[11] ZHAO X, ZHANG H, PAN X, et al. Fact-and-reflection (FaR) improves confidence calibration of large language models[C/OL]//KU L W, MARTINS A, SRIKUMAR V. Findings of the Association for Computational Linguistics: ACL 2024. Bangkok, Thailand: Association for Computational Linguistics, 2024: 8702-8718. https://aclanthology.org/2024.findings-acl.515/. DOI: 10.18653/v1/2024.findings-acl.515.

[12] KIM V T, KRUMDICK M, REDDY V, et al. An analysis of multilingual factscore[C/OL]//AL-ONAIZAN Y, BANSAL M, CHEN Y. Proceedings of the 2024 Conference on Empirical Methods in Natural Language Processing, EMNLP 2024, Miami, FL, USA, November 12-16, 2024. Association for Computational Linguistics, 2024: 4309-4333. https://aclanthology.org/2024.emnlp-main.247.

[13] RAWTE V, CHADHA A, SHETH A, et al. Tutorial proposal: Hallucination in large language models[C/OL]// KLINGER R, OKAZAKI N, CALZOLARI N, et al. Proceedings of the 2024 Joint International Conference on Computational Linguistics, Language Resources and Evaluation (LREC-COLING 2024): Tutorial Summaries. Torino, Italia: ELRA and ICCL, 2024: 68-72. https://aclanthology.org/2024.lrec-tutorials.11/.

[14] PARK S, DU X, YEH M H, et al. How to steer llm latents for hallucination detection?[A]. 2025.

[15] CHEN C, LIU K, CHEN Z, et al. INSIDE: llms’ internal states retain the power of hallucination detection [C/OL]//The Twelfth International Conference on Learning Representations, ICLR 2024, Vienna, Austria, May 7-11, 2024. OpenReview.net, 2024. https://openreview.net/forum?id=Zj12nzlQbz.

[16] GREENBLATT R, DENISON C, WRIGHT B, et al. Alignment faking in large language models[J/OL]. CoRR, 2024, abs/2412.14093. https://doi.org/10.48550/arXiv.2412.14093. DOI: 10.48550/ARXIV.2412.14093.

[17] HE J, GONG Y, LIN Z, et al. LLM factoscope: Uncovering LLMs’ factual discernment through measuring inner states [C/OL]//KU L W, MARTINS A, SRIKUMAR V. Findings of the Association for Computational Linguistics: ACL 2024. Bangkok, Thailand: Association for Computational Linguistics, 2024: 10218-10230. https://aclanthology.org/2024.findings-acl.608/. DOI: 10.18653/v1/2024.findings-acl.608.

[18] JU T, SUN W, DU W, et al. How large language models encode context knowledge? A layer-wise probing study[C/OL]//CALZOLARI N, KAN M, HOSTE V, et al. Proceedings of the 2024 Joint International Conference on Computational Linguistics, Language Resources and Evaluation, LREC/COLING 2024, 20-25 May, 2024, Torino, Italy. ELRA and ICCL, 2024: 8235-8246. https://aclanthology.org/2024.lrec-main.722.

[19] JIN M, YU Q, HUANG J, et al. Exploring concept depth: How large language models acquire knowledge and concept at different layers?[C/OL]//RAMBOW O, WANNER L, APIDIANAKI M, et al. Proceedings of the 31st International Conference on Computational Linguistics, COLING 2025, Abu Dhabi, UAE, January 19-24, 2025. Association for Computational Linguistics, 2025: 558-573. https://aclanthology.org/2025.coling-main.37/.

[20] HE L, CHEN P, NIE E, et al. Decoding probing: Revealing internal linguistic structures in neural language models using minimal pairs[C/OL]//CALZOLARI N, KAN M, HOSTE V, et al. Proceedings of the 2024 Joint International Conference on Computational Linguistics,

Language Resources and Evaluation, LREC/COLING 2024, 20-25 May, 2024, Torino, Italy. ELRA and ICCL, 2024: 4488-4497. https://aclanthology.org/2024.lrec-mai

n.402.

[21] BÜRGER L, HAMPRECHT F A, NADLER B. Truth is universal: Robust detection of lies in LLMs[C]//Advances in Neural Information Processing Systems: Vol. 37. 2024: 138393-138431.

[22] ZHANG S, YU T, FENG Y. TruthX: Alleviating hallucinations by editing large language models in truthful space[C/OL]//KU L W, MARTINS A, SRIKUMAR V. Proceedings of the 62nd Annual Meeting of the Association for Computational Linguistics (Volume 1: Long Papers). Bangkok, Thailand: Association for Computational Linguistics, 2024: 8908-8949. https://aclanthology.org/2024.acl-long.483/. DOI: 10.18653/v1/2024.acl-long.483.

[23] DU X, XIAO C, LI S. Haloscope: Harnessing unlabeled LLM generations for hallucination detection [C/OL]//GLOBERSONS A, MACKEY L, BELGRAVE D, et al. Advances in Neural Information Processing Systems 38: Annual Conference on Neural Information Processing Systems 2024, NeurIPS 2024, Vancouver, BC, Canada, December 10 - 15, 2024. 2024. http://papers.nips.cc/paper_files/paper/2024/hash/ba92705991cfbbcedc26e27e833ebbae-Abstract-Conference.html.

[24] SELLAM T, DAS D, PARIKH A. BLEURT: Learning robust metrics for text generation[C/OL]//JURAFSKY D, CHAI J, SCHLUTER N, et al. Proceedings of the 58th Annual Meeting of the Association for Computational Linguistics. Online: Association for Computational Linguistics, 2020: 7881-7892. https://aclanthology.org/2020.acl-main.704/. DOI: 10.18653/v1/2020.acl-main.704.